# SEMI-AUTOMATIC SIMULTANEOUS INTERPRETING QUALITY EVALUATION

Xiaojun Zhang

Division of Literature and Languages, University of Stirling, Stirling, UK

## ABSTRACT

*Increasing interpreting needs a more objective and automatic measurement. We hold a basic idea that 'translating means translating meaning' in that we can assessment interpretation quality by comparing the meaning of the interpreting output with the source input. That is, a translation unit of a 'chunk' named Frame which comes from frame semantics and its components named Frame Elements (FEs) which comes from Frame Net are proposed to explore their matching rate between target and source texts. A case study in this paper verifies the usability of semi-automatic graded semantic-scoring measurement for human simultaneous interpreting and shows how to use frame and FE matches to score. Experiments results show that the semantic-scoring metrics have a significantly correlation coefficient with human judgment.*

## 1. INTRODUCTION

With the increasing needs of international communication, the massive global interpreting demands rise in recent years. According to the survey of Common Sense Adversary, the portion of interpreting market has expanded to 14.68% (on-site interpreting 11.38%, telephone interpreting 2.22% and interpreting technology 1.08%) among the whole language service market in 2015 and the interpreting LSPs are sharing the total US$34,778 billion language service cake with others (DePalma et al., 2015). The huge interpreting demand may call for a more objective and automatic measurement to evaluate the quality of human interpreting.

The traditional quantitative metrics of interpreting quality (IQ) assessment is human scoring which is a professional evaluation by interpreting judges focusing on interpreters' performance in the booth such as fluency and adequacy of their translations, on-site response and interpreting skills. However, there is poor agreement on what constitutes an acceptable translation. Some judges regard a translation as unacceptable if a single word choice is suboptimal. At the other end of the scale, there are judges who will accept any translation that conveys the approximate meaning of the sentence, irrespective of how many grammatical or stylistic mistakes it contains. Without specifying more closely what is meant by "acceptable", it is difficult to compare evaluations.

The other manual assessment method is task-oriented evaluation which is an end-to-end user expectation test. It is the metrics that try to assess whether interpreting output is good enough that an audience is able to understand the content of the interpreted speech. In interpreting literature, maximum expectation (ME) is a normal-used standard to assess IQ which can meet the maximum requirement of audience to understand the speech content very well. This manual assessment is very vague, and it is difficult for evaluators to be consistent in their application. For example, audience A may get her/his maximum expectation from the English interpreting output "*Mary told the cake is to be cut by John.* " for s/he just is wondering "*Who will cut the cake?*" while audience B may get nothing for s/he is focusing on "*Could a cake be told?*".

The automatic evaluation metrics of IQ may be borrowed from spoken language translation (SLT) and machine translation (MT) evaluation. BLEU, METEOR, TER and other metrics can be introduced to measure human interpreting output. The problem is: these metrics are all based on references. And, there is no chance for interpreters to re-order their translation segments while they are doing their interpreting job simultaneously with the speaker. That is to say, it's unfair to evaluate human interpreting quality with machine translation metrics for interpreting itself is disfluent (Honel and Schultz, 2005). Speech translation is normally an interactive process, and it is natural that it should be less than completely automatic (Rayner et al., 2000). At a minimum, it is clearly reasonable in many contexts to feed back to the audience and permit him or her to abort translation if the interpreting output was unacceptably bad. Evaluation should take account of this possibility.

Generalized word posterior probability (GWPP) is often used as a key metric to confidentially estimate MT quality. Blatz et al. (2004) obtained the GWPP by calculating N-best list based on some statistical machine translation features such as semantic feature extracted from WordNet, shallow parsing feature and draw a conclusion that the estimation performance based on GWPP is better than that based on linguistic features. Ueffing et al. (2003, 2007) employed the methods of target position window, relative frequency and system models to count the GWPP and verify the efficiency of MT confidential estimation. Xiong et al. (2010) integrated GWPP with syntactic and lexical features into a binary classifier based on maximum entropy model to estimate the correctness and incorrectness of the word in MT hypothesis. Du and Wang (2013) tested different GWPP applications in MT estimation and showed a better performance when they combined the features of GWPP and linguistics. GWPP is a word-level confidential estimation metric. However, a meaningful chunk seems to be a better option to act as a translation unit in practice than a separate word. Specia et al. (2009) and Bach et al. (2011) adopted more features besides GWPP to estimate MT quality. Sakamoto et al. (2013) conducted a field test of a simultaneous interpreting system for face-to-face services to evaluate the 'solved task ratio' for tasks, which is a typical task-based evaluation. The above previous works are all focusing on MT evaluation but not human translation assessment Actually, there is still no real consensus on how to evaluate human translation including interpreting automatically. The most common approach is some version of the following: The system is run on a set of previously unseen data; the results are stored in text form; someone judges them as acceptable or unacceptable translations; and finally the system's performance is quoted as the proportion that is acceptable. Zhang (2010) introduced a fuzzy integrated evaluation method which applies fuzzy mathematics for multi-factor evaluation in human translation. Here, the computer was only a calculator to perform the acceptable proportion from human judges. Tian (2010) developed an online automatic scoring English-Chinese translation system, YanFa, which compares a translation output with reference by synonym matching, sentence-pattern matching and word similarity calculation respectively. Yanfa's semantic scoring has been explored on lexical level with resources of HowNet and Cilin. Both evaluation systems are concentrated on text translation. As to interpreting, there are some specific features of speech should be considered sufficiently.

Lo and Wu (2011) introduced a semi-automatic metric via semantic frames, MEANT, to assess translation utility by matching semantic role fillers, producing scores that correlate with human judgment as well as HTER but much lower labor cost. They (Lo and Wu, 2013) also contrasted the coefficiency on adequacy of semantic role labels (SRL) metric, syntactic based metric and n-gram based BLEU metric with human judgment and showed significantly more SRL's effective and better correlation with human judgment. SRL based metric also serves for MT but not human interpreting.

Here, we adopt the idea of semantic frame and semantic role based evaluation method and propose a translation unit of 'meaningful chunk' named 'frame' to explore the matched frames of target sentence from original sentence (not from reference), and a frame component of 'semantic role' named 'frame element (FE)' to explore the matched FEs in the target frames from the ones in the original frames as well. We introduce this semantic-scoring measurement into human interpreting domain and define two metrics: minimum expectation (MinE) for the frame matching to meet the minimum expectation of the audience to one's interpreting job and maximum expectation (MaxE) for the FE matching to meet the maximum expectation of the audience to one's interpreting job.

The aim of this paper is to verify the usability of automatic semantic-scoring measurement of human interpreting quality and show how to use frame and FE matches to score. Comparing with human scoring and automatic MT evaluation metrics, automatic semantic-scoring measurement has several distinctive features in interpreting quality assessment: (1) Scoring is based on meaning; (2) Scoring is reference-independent; (3) Scoring is graded to meet different requirements of audiences and (4) Less re-ordering operation in interpreting. Motivated by the above features, we make a pilot study on this issue oriented to computer-aided interpreting system development.

## 2. FRAME AND FRAME ELEMENT

A frame is a collection of facts that specify "characteristic features, attributes, and functions of a denotatum, and its characteristic interactions with things necessarily or typically associated with it." (Cruse, 2001) The semantic frame comes from frame semantics which extends Charles J. Fillmore's case grammar (Fillmore, 1968) from linguistic semantics to encyclopaedic knowledge. The basic idea is that "meanings are relativized to scenes" (Fillmore, 1977). That is, one cannot understand the meaning of a single word without access to all the essential knowledge that relates to that word. Thus, a word activates, or evokes, a frame of semantic knowledge relating to the specific concept it refers to. Frames are based on recurring experiences. The commercial transaction frame is based on recurring experiences of commercial transactions. Words not only highlight individual concepts, but also specify a certain perspective from which the frame is viewed. For example, "*sell*" views the situation from the perspective of the seller and "*buy*" from the perspective of the buyer. In this case the concept frame is applied to verbs like buy with the intention to represent the relationships between syntax and semantics (*Table 1*).

Table 1: An example of concept frame buy

| BUYER | buy | GOODS | (SELLER) | (PRICE) |
|---|---|---|---|---|
| subject | | object | from | for |
| Angela | bought | the owl | from Pete | for $ 10 |
| Eddy | bought | them | | for $ 1 |
| Penny | bought | a bicycle | from Stephen | |

Frame Net (http://framenet.icsi.berkeley.edu/fndrupal/home) is based on the theory of frame semantics (Baker, 2014). The basic idea is straightforward: that the meanings of most words can best be understood on the basis of a semantic frame: a description of a type of event, relation, or entity and the participants in it. For example, the concept of cooking typically involves a person doing the cooking (Cook), the food that is to be cooked (Food), something to hold the food while cooking (Container) and a source of heat (Heating_instrument). In the FrameNet project, this is represented as a frame called Apply_heat, and the Cook, Food, Heating_instrument and Container are called FEs . Words that evoke this frame, such as fry, bake, boil, and broil, are called lexical units (LUs) of the Apply_heat frame. The job of FrameNet is to define the frames and to annotate sentences to show how the FEs fit syntactically around the word that evokes the frame.

In FrameNet, the above sentence "*Mary told the cake to be cut by John.*" will be involved in two frames (*Figure 1*) and annotated as the following colorful markers of LUs (black background with upper case) and FEs (colourful background):

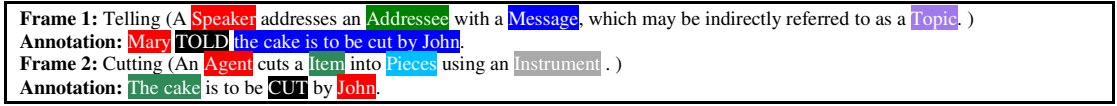

Figure 1: An example of Frame Net annotation

Our naïve idea of simultaneous interpreting quality measurement is that the more frames are matched with the source sentence the more meaning is transferred in the target interpreting sentence, and that the more FEs are matched with the source frames the more expectation is achieved to the target audience.

## 3. A CASE STUDY

### 3.1. Data and experiment

We choose a real-situation public speech, inaugural address by President Barack Obama in 2012, as the input source. The video, transcription and reference text translation of this inaugural are all acquired on line easily. The video of Chinese interpreting output by a senior interpreter (professional interpreter) is also obtained in open source. The senior interpreter works for Phoenix Television, a Hong Kong-based Mandarin and Cantonese-language television broadcast. We got her interpreting audio from Phoenix TV website (http://v.ifeng.com) and transcribed it manually. Our experiment begins from the frame and FE annotations of the transcriptions of Obama's speech and the senior interpreter's voice. We matched the frames and FEs of input and output and scored its MinE and MaxE. Then we scored other three outputs from junior interpreters (interpreting learners) and compared with the score of senior one to see the efficiency of this method. Also, we compare the scores of semantic-scoring and n-gram based BLEU of reference, different interpreters' interpreting outputs and MT output with their human scores to view their correlations.

### 3.2. Annotation

We annotated the semantic frame and their FEs manually for either there is no reliable semantic role annotator or the disfluent simultaneous interpreting is not suit for the state-of-art semantic annotator. As a pilot study we should grantee the gold standard of the testing data.

There are 87 sentences in total in the source text transcribed from the speech. We annotated these sentences which are similar to Frame Net annotation; for example, the source sentence 20 (SS20) can be annotated as *Figure 2*. In the following annotation examples, LUs are highlighted and FEs are bracketed and marked.

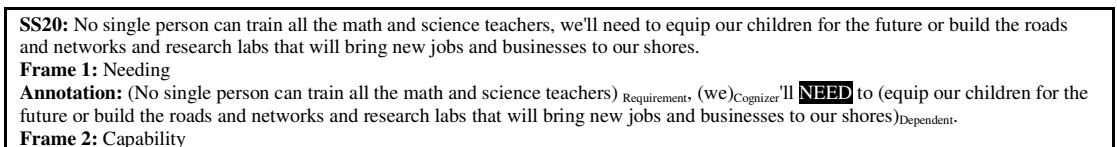

**Annotation:** (No single person)<sub>Entity</sub> CAN (train all the math and science teachers)<sub>Event</sub>
**Frame 3:** Education_teaching
**Annotation:** TRAIN (all the math and science teachers)<sub>Fact</sub>
**Frame 4:** Supply
**Annotation:** EQUIP (our children)<sub>Recipient</sub> (for the future)<sub>Purpose_of_recipient</sub>
**Frame 5:** Building
**Annotation:** BUILD (the roads)<sub>Created_entity</sub> and (networks)<sub>Created_entity</sub> and (research labs)<sub>Created_entity</sub> that will bring new jobs and businesses to our shores.
**Frame 6:** Bringing
**Annotation:** (that)<sub>Agent</sub> will BRING (new jobs)<sub>Theme</sub> and (businesses)<sub>Theme</sub> (to our shores)<sub>Goal</sub>

Figure 2: SS20 annotation

And, the senior interpreter's output of sentence 20 (SI20) can be annotated as Figure 3:

**SI20:**没有任何一个人有能力训练出我们后代的教育需要的所有数学和科学教师，或者建造出能把新的工作和商业机会带给我们的道路、网络、实验室。
**Frame 1:** Capability
**Annotation:** (没有任何一个人)<sub>Entity</sub> 有能力 (训练出我们后代的教育需要的所有数学和科学教师)<sub>Event</sub>，或者(建造出能把新的工作和商业机会带给我们的道路、网络、实验室)<sub>Event</sub>。
**Frame 2:** Existence
**Annotation:** 没有(任何一个人)<sub>Entity</sub>
**Frame 3:** Education_teaching
**Annotation:** 训练出(我们后代的教育需要的所有数学和科学教师)<sub>Fact</sub>
**Frame 4:** Needing
**Annotation:** (我们后代的教育)<sub>Cognizer</sub> 需要
**Frame 5:** Building
**Annotation:** 建造出能把新的工作和商业机会带给我们的(道路)<sub>Created_entity</sub>、(网络)<sub>Created_entity</sub>、(实验室)<sub>Created_entity</sub>
**Frame 6:** Bringing
**Annotation:** 把(新的工作)<sub>Theme</sub>和(商业机会)<sub>Theme</sub>带给(我们)<sub>Goal</sub>

Figure 3: SI20 annotation

For the first junior interpreter's output of sentence 20 (JI$_{01}$20), we annotated it as Figure 4:

**JI$_{01}$20:** 没有一个单独的人可以培训美国的科学家，可以保护美国孩子们的未来，不可能把我们的工作保护在我们的国土内。
**Frame 1:** Capability
**Annotation:** (没有一个单独的人)<sub>Entity</sub> 可以 (培训美国的科学家)<sub>Event</sub>，可以 (保护美国孩子们的未来)<sub>Event</sub>，不可能 (把我们的工作保护在我们的国土内)<sub>Event</sub>。
**Frame 2:** Existence
**Annotation:** 没有 (一个单独的人)<sub>Entity</sub>
**Frame 3:** Education_teaching
**Annotation:** 培训 (美国的科学家)<sub>Student</sub>
**Frame 4:** Protecting
**Annotation:** 保护 (美国孩子们的未来)<sub>Asset</sub>
**Frame 5:** Protecting
**Annotation:** 把(我们的工作)<sub>Asset</sub> 保护 在我们的国土内)<sub>Place</sub>

Figure 4: JI$_{01}$20 annotation

JI$_{02}$20 and JI$_{03}$20 are annotated as Figure 5 and Figure 6:

**JI$_{02}$20:** 没有一个单独的人能够训练所有的数学和科学家。我们需要为将来装备我们的孩子和实验室，这将带来新的工作和商业。
**Frame 1:** Capability

**Annotation:** (没有一个单独的人)$_{Entity}$ 能够 (训练所有的数学和科学家)$_{Event}$
**Frame 2:** Existence
**Annotation:** 没有 (一个单独的人)$_{Entity}$
**Frame 3:** Education_teaching
**Annotation:** 培练 (所有的数学和科学家)$_{Student}$
**Frame 4:** Needing
**Annotation:** (我们)$_{Cognizer}$ 需要 (为将来装备我们的孩子和实验室)$_{Dependent}$
**Frame 5:** Supply
**Annotation:** (为将来)$_{Purpose\_of\_Recipient}$ 装备 (我们的孩子和实验室)$_{Recipient}$
**Frame 6:** Bringing
**Annotation:** (这)$_{Agent}$ 将带来 (新的工作和商业)$_{Theme}$

Figure 5: $JI_{02}20$ annotation

**$JI_{03}20$:** 没有一个人能够训练科学家。我们需要增强孩子，修建实验室，这将给我们的海岸带来新的工作机会。
**Frame 1:** Capability
**Annotation:** (没有一个人)$_{Entity}$ 能够 (训练科学家)$_{Event}$
**Frame 2:** Existence
**Annotation:** 没有 (一个人)$_{Entity}$
**Frame 3:** Education_teaching
**Annotation:** 培训 (科学家)$_{Student}$
**Frame 4:** Needing
**Annotation:** (我们)$_{Cognizer}$ 需要 (增强孩子)$_{Dependent}$，(修建实验室)$_{Dependent}$
**Frame 5:** Cause_change_of_strenghth
**Annotation:** 增强 (孩子)$_{Patient}$
**Frame 6:** Building
**Annotation:** 修建 (实验室)$_{Create\_entity}$
**Frame 7:** Bringing
**Annotation:** (这)$_{Agent}$ 将给 (我们的海岸)$_{Goal}$ 带来 (新的工作机会)$_{Theme}$

Figure 6: $JI_{03}20$ annotation

## 3.3. Evaluation metrics

We generate the annotation list for each source sentence and its interpreting outputs of different interpreters. Take sentence 20 as an example, the annotation lists are in Table 2:

Table 2: Annotation list of sentence 20

| | Frame | Frame Element (FE) |
|---|---|---|
| SS20 | Needing | Requirement, Cognizer, Dependent |
| | Capability | Entity, Event |
| | Education_teaching | Fact |
| | Supply | Recipient, Purpose_of_recipient |
| | Building | Created_entity, Created_entity, Created_entity |
| | Bringing | Agent, Theme, Theme, Goal |
| SI20 | Capability | Entity, Event, Event |
| | Existence | Entity |
| | Education_teaching | Fact |
| | Needing | Cognizer |
| | Building | Created_entity, Created_entity, Created_entity |
| | Bringing | Theme, Theme, Goal |
| $JI_{01}20$ | Capability | Entity, Event, Event, Event |
| | Existence | Entity |
| | Education_teaching | Student |
| | Protecting | Asset |
| | Protecting | Asset, Place |
| $JI_{02}20$ | Capability | Entity, Event |

| | Existence | Entity |
|---|---|---|
| | Education_teaching | Student |
| | Needing | Cognizer, Dependent |
| | Supply | Purpose_of_Recipient, Recipient |
| | Bringing | Agent, Theme |
| JI₀₃20 | Capability | Entity, Event |
| | Existence | Entity |
| | Education_teaching | Student |
| | Needing | Cognizer, Dependent, Dependent |
| | Cause_of_strength | Patient |
| | Building | Create_entity |
| | Bringing | Agent, Goal, Theme |

We set two metrics to score the interpreting outputs: MinE and MaxE. MinE refers to the expectation value which can meet the audiences' 'basic' requirement, that is, the frames of output can match the most frames of speaker's input. MaxE refers to the expectation value which can meet the audiences' 'advanced' requirement, that is, the FEs of output frames can match the most FEs of speaker's input frames. To qualify the above metrics, we define the both in terms of F-score that balance each precision and recall analysis. Precision of MinE ($P_{MinE}$) counts the number of matched frames ($N_m$)in the total frames of the target sentence ($N_t$), as

$$P_{MinE} = \frac{N_m}{N_t}$$

(1)

Precision of MaxE ($P_{MaxE}$) counts the number of matched FEs in the matched frames ($n_m$)in the total FEs of the target sentence frames ($n_t$), as

$$P_{MaxE} = \sum_{matched} \frac{n_m}{n_t}$$

(2)

Recall of MinE ($R_{MinE}$) counts the number of matched frames ($N_m$)in the total frames of the source sentence ($N_s$), as

$$R_{MinE} = \frac{N_m}{N_s}$$

(3)

Recall of MaxE ($R_{MaxE}$) counts the number of matched FEs in the matched frames ($n_m$)in the total FEs of the source sentence frames ($n_s$), as

$$R_{MaxE} = \sum_{matched} \frac{n_m}{n_s}$$

(4)

F-score of MinE ($F_{MinE}$) is the balance of precision of MinE and recall of MinE, as

$$F_{MinE} = \frac{2 \times P_{MinE} \times R_{MinE}}{P_{MinE} + R_{MinE}}$$

(5)

F-score of MaxE ($F_{MaxE}$) is the balance of precision of MaxE and recall of MaxE, as

$$F_{MaxE} = \frac{2 \times P_{MaxE} \times R_{MaxE}}{P_{MaxE} + R_{MaxE}}$$

(6)

In the example of SI20, the number of senior interpreter's matched frames ($N_m$) is 5, there are 6 frames in the target sentence ($N_t$), so for this sentence the precision of MinE ($P_{MinE}$) is 5/6; there are 6 frames of the source sentence ($N_s$), so recall of MinE ($R_{MinE}$) of this sentence to the senior interpreter is 5/6, and its F-score ($F_{MinE}$) is 0.83. Accordingly, in this example, the number of matched FEs in the matched frames ($n_m$) is 10, there are 12 FEs in the target sentence frames ($n_t$), so, its precision of MaxE ($P_{MaxE}$) is 10/12; there are 15 FEs in the source sentence frames ($n_s$), so recall of MaxE ($R_{MaxE}$) of this sentence to the senior interpreter is 10/15, and its F-score ($F_{MaxE}$) is 0.74.

Table 3: Scores of MinE and MaxE of sentence 20

|  | SI20 | JI$_{01}$20 | JI$_{02}$20 | JI$_{03}$20 |
|---|---|---|---|---|
| $P_{MinE}$ | 0.83 | 0.40 | 0.83 | 0.71 |
| $R_{MinE}$ | 0.83 | 0.33 | 0.83 | 0.83 |
| $F_{MinE}$ | 0.83 | 0.36 | 0.83 | 0.77 |
| $P_{MaxE}$ | 0.83 | 0.22 | 0.80 | 0.66 |
| $R_{MaxE}$ | 0.67 | 0.13 | 0.53 | 0.53 |
| $F_{MaxE}$ | 0.74 | 0.16 | 0.64 | 0.59 |

Hence, we obtain $P_{MinE}$, $R_{MinE}$, $F_{MinE}$, $P_{MaxE}$, $R_{MaxE}$ and $F_{MaxE}$ of this sentence to senior interpreter and three junior interpreters respectively as *Table 3*. *Table 4* is the average sentence-level F-scores of MinE and MaxE for the four interpreting outputs:

Table 4: The average F-scores of MinE and MaxE

|  | SI | JI$_{01}$ | JI$_{02}$ | JI$_{03}$ |
|---|---|---|---|---|
| $F_{MinE}$ | 0.82 | 0.41 | 0.78 | 0.73 |
| $F_{MaxE}$ | 0.75 | 0.25 | 0.66 | 0.54 |

There are two issues in the metric MinE: one is the order of the matched frames pairs, and the other is the similarity of matched frames. The former is not a real issue for the interpreters has the feasibility to deal with their output in different order. The latter should relate to the second metric, MaxE.

It seems so cruel to the matched FEs in the unmatched frames that they will be counted zero in the formula. Actually that's the unbalanced points in human scoring. For example, in the frame of 'Existence' of SI20 and JI$_{01}$20, the FE 'Entity' matches the one in the frame of 'Capability' of SS20. Unfortunately, the chunk "*No single person*" is not meaningful in the source language and it is not a dependent frame even though its translation in the target language, "没有任何一个人", can be annotated as the frame 'Existence'.

To the interpreting outputs, there are other two issues in the metric MaxE: one is the repetitive or added FEs in one frame, and the other is keywords mistranslation. As we mentioned that interpreting is disfluent, repetitive expressions are always occurred in one's interpreting, as well as some missing expressions. The missing expressions mean the missing FEs to reduce the MaxE value in scoring; while to the repetitive ones, or sometimes due to the experienced processing by senior interpreters, the added expresses which can make output more smooth and fluent will add the number of FEs in the target fram. For example, in the sentence 12, annotations as *Figure 7*:

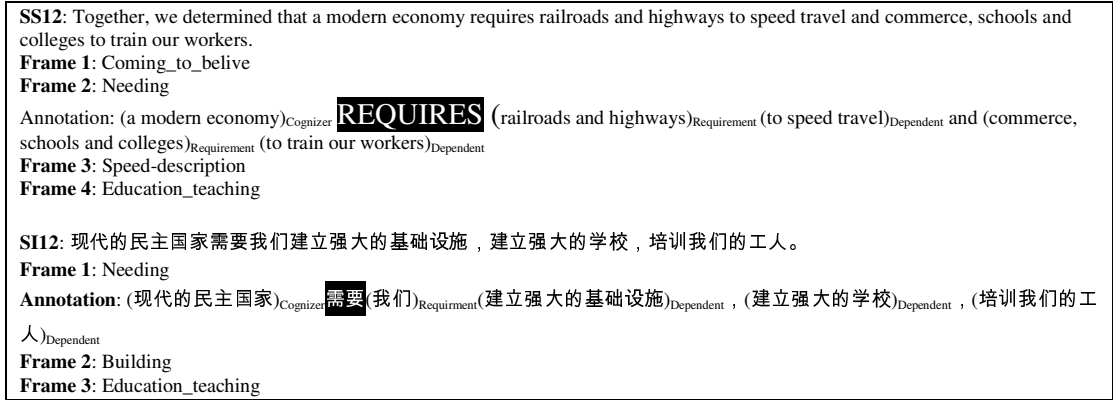

Figure 7: Annotations of sentence 12

In the matched frame 'Needing', there are 5 FEs in each of the original frame and target frame. But the FE 'Dependent' in target frame occurs 3 times while twice in the original sentence in that the interpreter made some smoothening treatment in his/her interpreting, which makes a trouble for us to count recall of this FE (3/2?). In this case, we normalized that recall in the range of [0,1]. That means, the matched FE, 'Dependent', in the above example, will be count only once by cutting off 1 FE in the target frame.

Distinguished with text translation, interpreting quality depends on the meaning transferring of several keywords in the speech. Terminologies, human or place names, time expressions, numbers (values, figures, amounts...) are the main four categories keywords. If these keywords were misunderstood by the audience, the interpreting is failure, even though their FEs were matched with the original frame. Hence, we set a penalty in scoring, that is, if one keyword was mistranslated, its matched FE number will reduce 1. For example, in *Figure 8*:

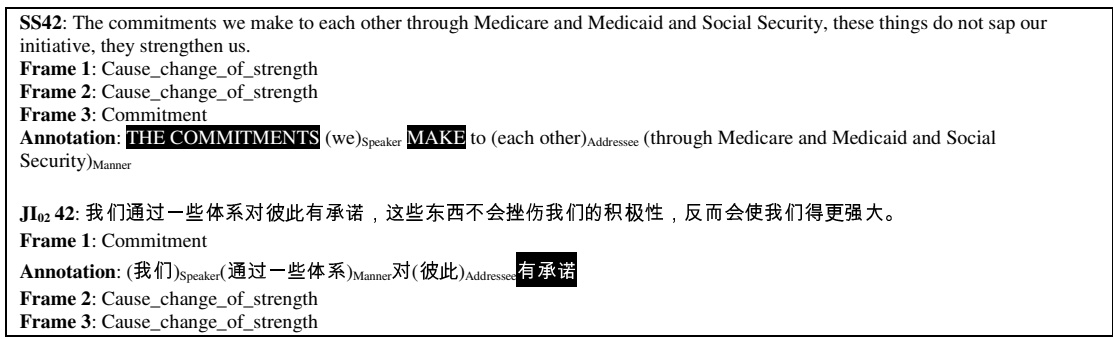

Figure 8: Annotations of sentence 42

In the matched frame 'Commitment', the source and the target frame share the same FE, 'Manner' and the output seems perfect. But the target 'Manner', "一些体系 (some systems)" , is more general in meaning than the source 'Manner', "*Medicare and Medicaid and Social Security*", which are all keywords in Obama's speech. In our scoring, this FE is not matched.

### 3.4. Correlation with human scoring

We use the Spearman's rank correlation coefficient ρ to measure the correlation of the semantic-scoring metrics with the human judgement and n-gram-based BLEU evaluation at sentence-level and took average on the whole speech interpreting data. We invite a professional interpreting teacher (English to Chinese) as a judge to score all the outputs from reference, senior interpreter, three junior interpreters and MT. We obtained the reference from an open-source language-learning website (http://news.iciba.com/study/listen/1554958.shtml) and MT output from Google Translate. The Spearman's rank correlation coefficient ρ can be calculated using the following equation:

$$\rho = 1 - \frac{6\sum d_i^2}{n(n^2 - 1)} \qquad (7)$$

where $d_i$ is the difference between the ranks of the evaluation metrics and the human judgement over of system $i$ and $n$ is the number of systems. The range of possible values of correlation coefficient is [-1,1], where 1 means the systems are ranked in the same order as the human judgement and -1 means the systems are ranked in the reverse order as the human judgment. The higher the value for ρ indicates the more similar the ranking by the evaluation metric to the human judgment.

*Table 5* shows the raw scores of example sentence 20 under our semantic-scoring metrics, MinE and MaxE, sentence-level BLEU and human scores with the corresponding ranks assigned to each of the systems.

Table 5: MinE/MaxE vs. BLEU in correlation with human judgment of example sentence 20

|  |  | Reference20 | SI20 | MT20 |
|---|---|---|---|---|
| $F_{MinE}$ | Score | 0.85 | 0.83 | 0.77 |
|  | Rank | 1 | 2 | 3 |
| $F_{MaxE}$ | Score | 0.80 | 0.74 | 0.59 |
|  | Rank | 1 | 2 | 3 |
| BLEU | Score | 1.00 | 0.13 | 0.14 |
|  | Rank | 1 | 3 | 2 |
| Human scoring | Score | 90 | 80 | 60 |
|  | Rank | 1 | 2 | 3 |

Our results show that the proposed semantic-scoring metrics have higher correlation with the human judgement than BLEU metric. *Table 6* compares the average sentence-level ρ for our proposed MinE, MaxE with BLEU. The correlation coefficient for MinE metric is 0.71 and MaxE is 0.73, while that for BLEU is 0.24. Our proposed metrics is significantly better than BLEU.

Table 6: Average sentence-level correlation for the evaluation metrics

| Metric | Correlation with human judgment |
|---|---|
| MinE | 0.71 |
| MaxE | 0.83 |
| BLEU | 0.24 |

## 5. CONCLUSIONS

"Translating means translating meaning. (Nida, 1986) "Comparing with text translation, interpreting has less distinctive features such as genre (literature or technical?), culture (domestication or foreignization?), sociological factors (post-colony or femelism?). The only thing what interpreters are focusing on is transferring the meaning of speaker's speech to the audience. On language structure, interpreting, especially, simultaneous interpreting allows little time for the interpreters in the booth to organize their outputs in the normal way as the target languages does. They always output their translation as the source language structure. So, the meaning-focused and less re-ordering give a possibility to evaluate the interpreting automatically and semantically.

Interpreting quality evaluation or estimation is an important part to the study of computer-aided interpreting study. We propose the semantic-scoring evaluation metrics which shows a significant advantage in correlation coefficient with human judgment. As a case study, we satisfied the experiment results. However, it is still a long way to reach our goal of fully automatically semantic-scoring interpreting outputs.

# ACKNOWLEDGEMENTS

This work is partly supported by Centre of Translation Studies of GDUFS project (Grant No.: CTS201501) and the Open Project Program of National Laboratory of Pattern Recognition (Grant No.: 201407353).

**Authors**

Dr. Xiaojun Zhang is a lecturer in translation and interpreting studies at Division of Literature and Languages, Faculty of Art and Humanities in University of Stirling. His researches include translation technologies such as computer-aided translation (CAT) and machine translation (MT), and translation and interpreting studies from their cognitive and psycholinguistic aspects.

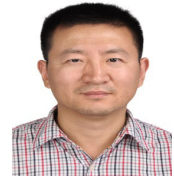